\documentclass[runningheads]{llncs}
\usepackage[T1]{fontenc}
\usepackage{graphicx}
\graphicspath{{./figs/}}
\usepackage{float} % JKB

\usepackage[caption=false]{subfig}
\begin{document}
\title{Modular Graph Extraction for Handwritten Circuit Diagram Images}
\author{Johannes Bayer\inst{1,2}\orcidID{0000-0002-0728-8735} \and
Leo van Waveren\inst{1}\orcidID{0000-0002-3817-8109} \and
Andreas Dengel\inst{1,2}\orcidID{0000-0002-6100-8255}}

\authorrunning{J. Bayer et al.}
\institute{RPTU, Erwin-Schrödinger-Straße 52, 67663 Kaiserslautern, Germany \and
           DFKI, Trippstadter Str. 17, 67663 Kaiserslautern, Germany
  \email{\{johannes.bayer,andreas.dengel\}@dfki.com}\\
  \url{https://fd-tech.informatik.uni-kl.de/}\\
  \email{\{jbayer,leo.vanwaveren\}@rptu.de}\\
  \url{https://www.dfki.de/}}
\maketitle              % typeset the header of the contribution
\begin{abstract}
  As digitization in engineering progressed, circuit diagrams (also referred to as schematics) are typically developed and maintained in computer-aided engineering (CAE) systems, thus allowing for automated verification, simulation and further processing in downstream engineering steps. However, apart from printed legacy schematics, hand-drawn circuit diagrams are still used today in the educational domain, where they serve as an easily accessible mean for trainees and students to learn drawing this type of diagrams. Furthermore, hand-drawn schematics are typically used in examinations due to legal constraints. In order to harness the capabilities of digital circuit representations, automated means for extracting the electrical graph from raster graphics are required.

  While respective approaches have been proposed in literature, they are typically conducted on small or non-disclosed datasets. This paper describes a modular end-to-end solution on a larger, public dataset, in which approaches for the individual sub-tasks are evaluated to form a new baseline. These sub-tasks include object detection (for electrical symbols and texts), binary segmentation (drafter's stroke vs. background), handwritten character recognition and orientation regression for electrical symbols and texts. Furthermore, computer-vision graph assembly and rectification algorithms are presented. All methods are integrated in a publicly available prototype.

\end{abstract}

\section{Introduction}
Circuit diagrams are symbolic, graph-based descriptions of electric and electronic structures. Apart from their usage in development processes (which typically take place in a digitized manner nowadays), they are also utilized in educational contexts like teaching and examinations. Often, these contexts require analogue, pen and paper-based information capturing. In order to support teachers and students with the existing, automated verification and simulation tools, digitization provides opportunities to enhance timely and objective feedback. To automate this process, an end-to-end solution is proposed which takes a raster graphic as input and creates a representation readable by computer-aided engineering (CAE) programs. The described approach assumes an image taken under real-world conditions, i.e. a snapshot taken by a mobile camera rather than a professional scanning device. These conditions are assumed to be more challening to evaluate than i.e. clean scans of printed schematics or rendered vector graphics.

\section{Related Work}
Generally, the existing literature on handwritten circuit diagram extraction lacks evaluation on large public datasets, preventing comparisons between approaches. Some authors even acknowledge the scarcity of those resources and create their own datasets. For example, \cite{rachala2022hand} states that ``Since there is no publicly available dataset, a custom dataset is generated by collecting hand-drawn circuits drawn by five different people.'' (working with a dataset that contains $5$ classes of electrical symbols in $154$ circuit images only). \cite{rabbani2016hand} utilizes a slightly larger dataset for their MLP-based symbol recognition approach, but also don't disclose their dataset. \cite{lakshman2019handwritten} also disclose very little information about the dataset they used for their approach based on local binary pattern and finite state machines.

\subsection{The CGHD Dataset}
CGHD\cite{thoma2021public} has been created as a public\footnote{\url{https://zenodo.org/records/10056817}} ground truth dataset of handwritten circuit diagram images. Over the last years, it has been extened \cite{bayer2023text,icpram23} in size, quality and features. The key metrics of the dataset are given below (see tab.~\ref{tab:cghd_features}).

\begin{table}
  \caption{CGHD Key Features.}
  \centering
  \label{tab:cghd_features}
    \begin{tabular}{|l|r|}
      \hline
      Feature           &  Count\\
      \hline
      Raw Images        & 2.424 \\
      Bounding Boxes    & 201.142\\
      Text Strings      & 61.595\\
      Segmentation Maps & 257\\
      Polygons          & 19.562\\
      Object Classes    & 59\\
      \hline
  \end{tabular}
\end{table}

\subsubsection{Images}
$25$ Drafters contributed to the dataset under controlled circumstances, additional circuits from real-world examinations provided by external partners are also added. Each of the drafters was asked to draw $12$ circuits, each of them twice, and to take $4$ images of each drawing (using different camera positions, illuminations and paper degradations), resulting in more than $300$ circuits covered. Different pencil and media types have been used, including plain, lined and squared paper as well as whiteboard and digitizer capturings. The schematics in CGHD feature a wide range of complexity from simple resistor circuits to complex radio and digital logic. The images of CGHD are intended to capture smartphone snapshots without further processing (few logos have been removed for copyright reasons).

\subsubsection{Bounding Box Annotations}
The annotated regions of interest (RoI) include electrical symbols, texts as well as structural features like junction points, wire corners and hops are described as bounding boxes (BB) along with their class.

\subsubsection{Symbol and Text Orientation}
In order to recover connector positions of electrical symbols and to aid the text recognition, the orientation of both is annotated. For electrical symbols, their rotation with respect to a symbol template is used (resolution 1 degree). For texts, 90 degree rotations are used as a mean to minimize annotation overhead.

\subsubsection{Text Strings}
Text BBs are annotated with their content as human-readable string. It is worth noting that apart from mostly latin and number characters, domain-specific special characters like $\mu$ and $\Omega$ are present. 

\subsubsection{Segmentation Maps}
Binary segmentation map have been added to tell drafter's strokes apart from background. They are intended to address complicated illumination, surrounding coffee-table objects as well as lines and squares of the paper background.

\subsubsection{Polygon Annotation}
\begin{figure}
  \centering
  \subfloat[Raw Image]{%
  \includegraphics[width=0.49\textwidth]{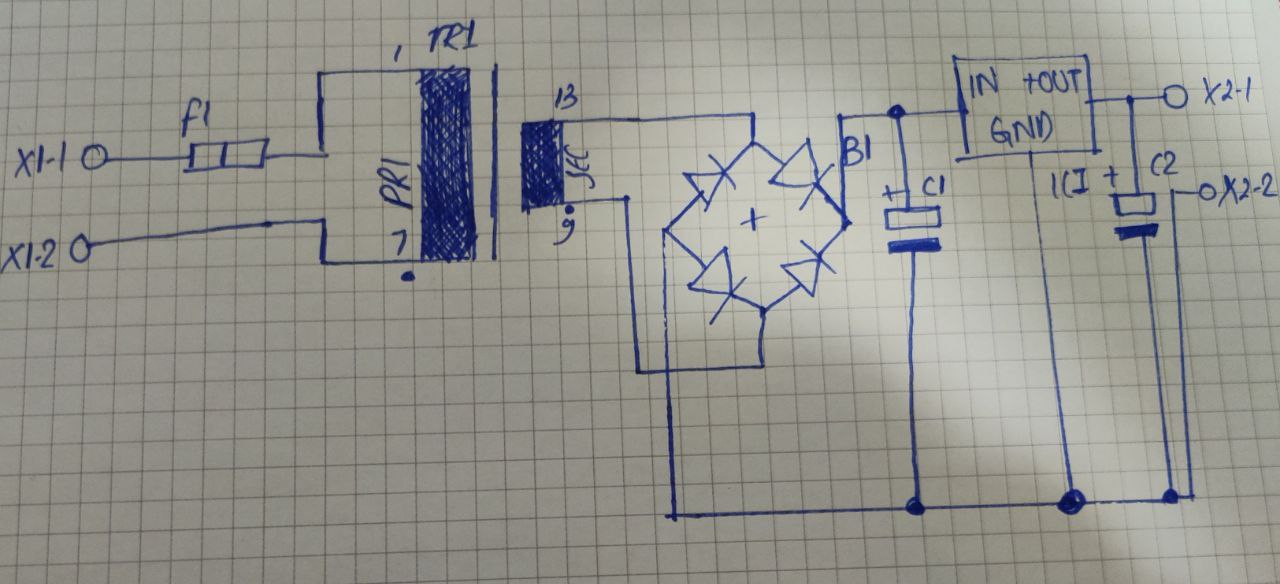}%
  }
  \hfil
  \subfloat[Binary Segmentation Map]{%
  \includegraphics[width=0.49\textwidth]{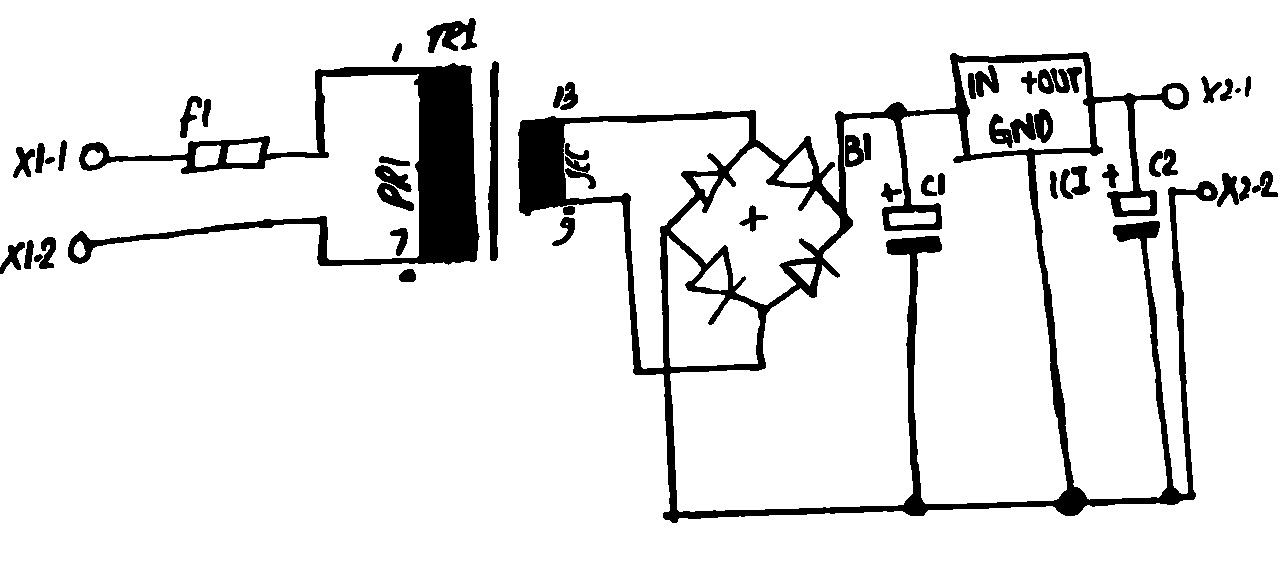}%
  }
  \hfil
  \subfloat[Polygon Annotations]{%
  \includegraphics[width=0.49\textwidth]{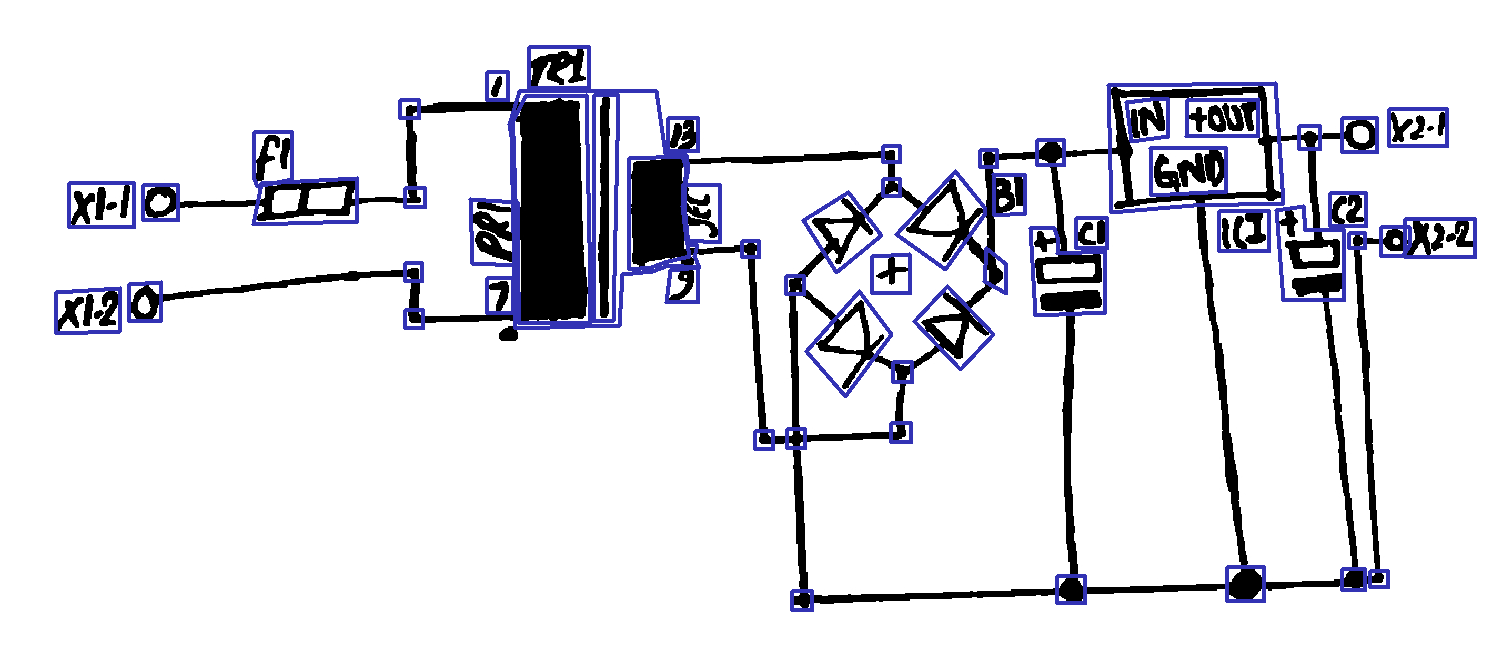}%
  }
  \hfil
  \subfloat[Semantic Segmentation Map]{%
  \includegraphics[width=0.49\textwidth]{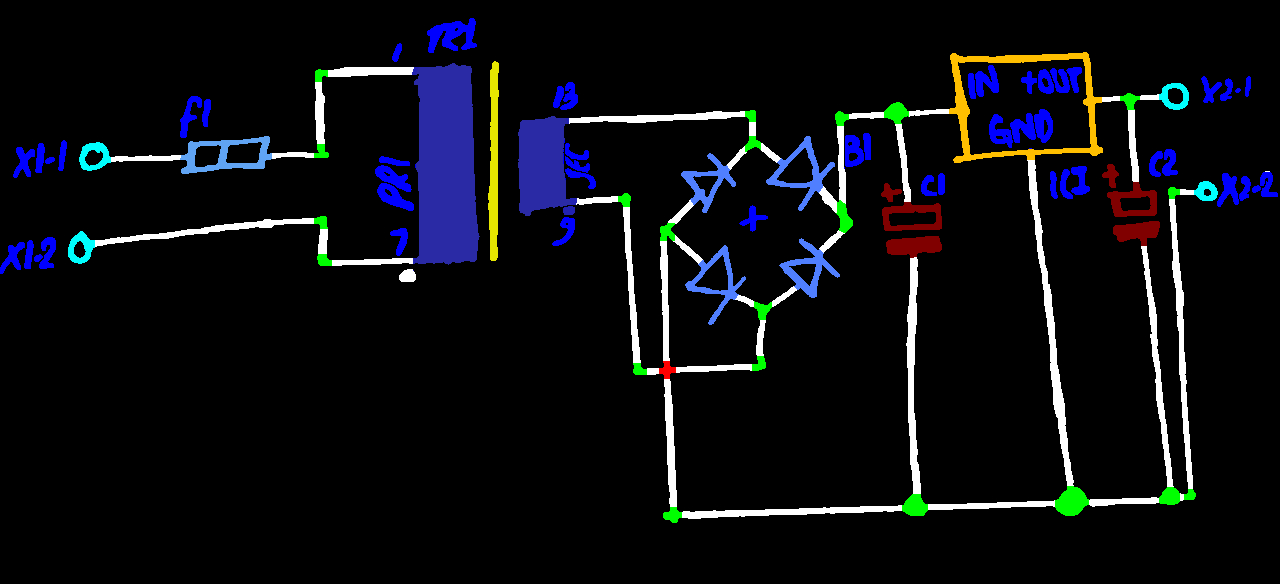}%
  }
  \caption{A semantic segmentation map can be synthesized from individual parts of the CGHD dataset.}
  \label{fig:pipeline}
\end{figure}

Based on the segmentation maps and the BB annotations, polygon annotations are created in a semi-automated fashion to avoid overlaps between multiple BBs. Scripts for refining the polygon annotations based on the actual segmentation map content with pixel-level accuracy while automatically creating (not manually annotated) wire polygons are shipped with the dataset (see fig.~\ref{fig:pipeline}).

\subsubsection{User Interface}
\begin{figure}
  \includegraphics[width=\textwidth]{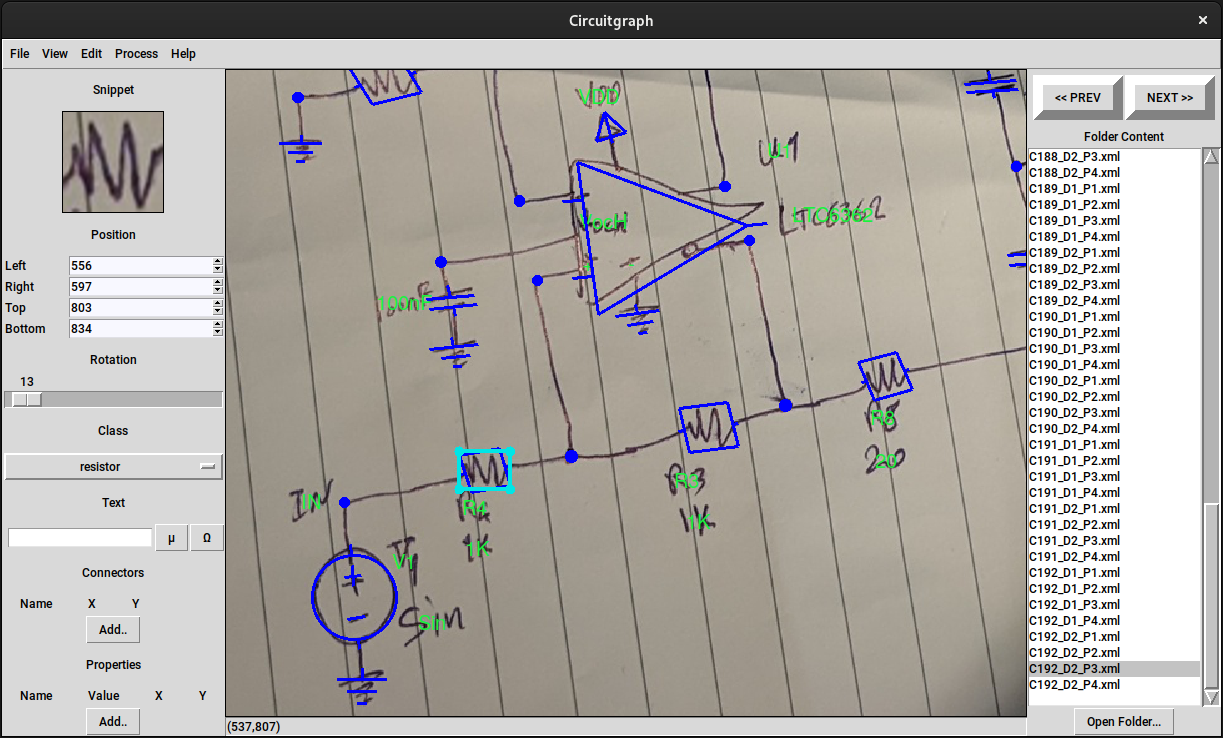}
  \caption{Circuitgraph User Interface.}
  \label{fig:ui}
\end{figure}

In order to edit customized annotations (like symbol rotations) conveniently, a respective user interface was created (see fig.~\ref{fig:ui}) and made publicly available.\footnote{\url{https://gitlab.com/circuitgraph}}

\section{Methodology}
In order to have a fully modular end-to-end solution for graph structure extraction from raster graphics, the individual steps are solved by independent models and algorithms:

\subsubsection{Object detection}
is applied to the input image, yielding the RoI's BB positions and classes. A Faster R-CNN\cite{ren2015faster} is used for this purpose.

\subsubsection{Binary segmentation}
is performed using a U-Net \cite{ronneberger2015u}. A fixed patch size is used for both training and inference. During inference, the image is cut into adjacent tiles, each equaling to the patch size. The resuling gap between the image's width and height and the multitude of the patch size fitting in the image is closed by performing the tiling process multiple times, each starting at different positions in the image(left-top, right-top, bottom-left).

\subsubsection{Orientation Regression}
\begin{figure}%[htp]
  \centering
  \subfloat[Resistor]{%
  \includegraphics[width=0.1\textwidth]{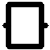}%
  }
  \hfil
  \subfloat[Capacitor]{%
  \includegraphics[width=0.1\textwidth]{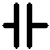}%
  }
  \hfil
  \subfloat[Diode]{%
  \includegraphics[width=0.1\textwidth]{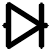}%
  }
  \hfil
  \subfloat[Ground]{%
  \includegraphics[width=0.1\textwidth]{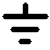}%
  }
  \hfil
  \subfloat[Logical And]{%
  \includegraphics[width=0.1\textwidth]{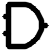}%
  }
  \hfil
  \subfloat[Lamp]{%
  \includegraphics[width=0.1\textwidth]{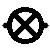}%
  }
  \hfil
  \subfloat[Speaker]{%
  \includegraphics[width=0.1\textwidth]{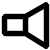}%
  }
  \caption{Selected Symbol Templates used as Reference for Orientation Prediction.}
  \label{fig:symbols}
\end{figure}

is performed on the image snippet for all objects of text and electrical symbol classes. A CNN-MLP model is used here, in which both the snippet and the respective symbol template (see fig.~\ref{fig:symbols}) are presented as input (for texts, the input is a zero-filled tensor). The output of this model is the sine and cosine encoded smallest possible angle between template and snippet. Since some symbols are mirror-symmetric (e.g. resistors) while orientation annotations are generally defined between $1$ and $360$ degree, the annotations of the respective classes require preprocessing to limit their range.

\subsubsection{Text Recognition}
is performed on the image snippet for all objects of class text. An LSTM\cite{hochreiter1997long} is used here.

\subsubsection{Edge Extraction}
is performed on the binary segmentation map in which all object regions have been removed. Since these object regions also include structural elements like line corners and crossings, the remaing image is expected to contain line segments only. Blob detection is applied to infer the positon of wire lines. For each blob wich touches two objects, an edge is created along with point-positioned ports in the respective nodes.

\subsubsection{Post-Processing}
involves the resolution of wire hops and rectification of uneven edge lines. Furthermore, the ports are corrected in position and assigned names from a symbol library.

After training the individual models, the end-to-end process is evaluated on a test set sample.

\section{Experiments}
From the CGHD dataset $21$ drafters are used as training (drafter $1$-$20$, $25$), $2$ drafters are used for validation (drafter $21$ and $22$) and $2$ drafters are for test (drafter $23$ and $24$).

\subsection{Object Detection}
\begin{figure}
  \centering
  \includegraphics[width=0.7\textwidth]{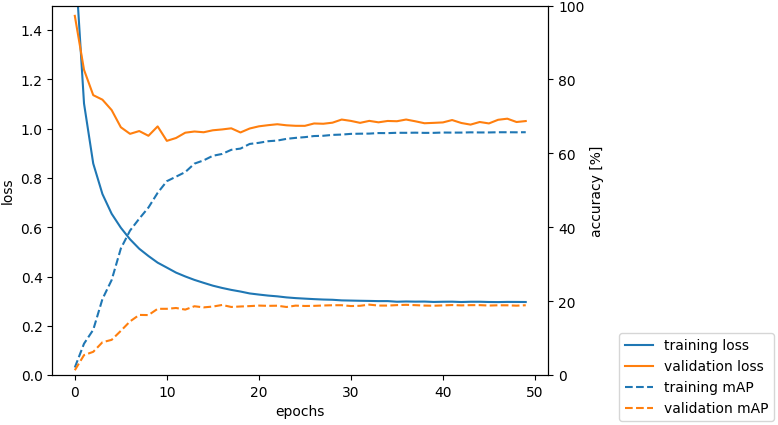}
  \caption{Learning Curve for Object Detection}
  \label{fig:learning_curve_object_detection}
\end{figure}

A standard Faster-RCNN from the torchvision implementation with a ResNet\cite{he2016deep}-152 backbone was used for object recognition. This model was trained using as SGD with a learning rate of $0.01$, a weight decay of $0.0005$ and a momentum of $0.9$. It yielded an mAP accuracy on the validation set of $18\%$ (see fig.~\ref{fig:learning_curve_object_detection}).

\subsection{Orientation Regression}
\begin{figure}[htp]
  \centering
  \subfloat[Snippet Input Only]{%
  \includegraphics[width=0.5\textwidth]{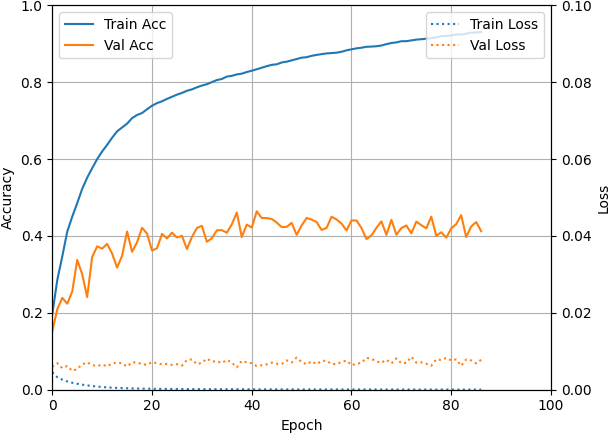}%
  }
  \hfil
  \subfloat[Template-Assisted]{%
  \includegraphics[width=0.5\textwidth]{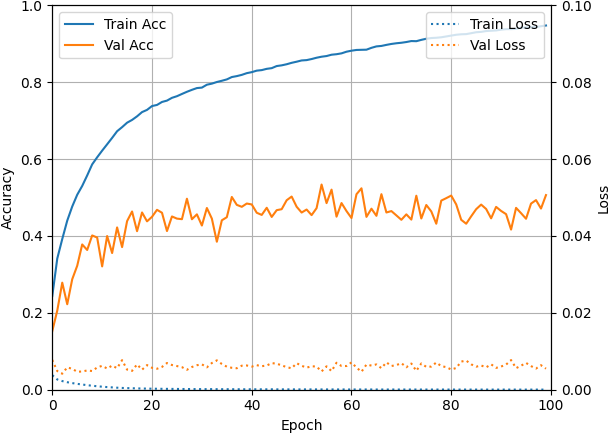}%
  }
  \caption{Learning Curves for Orientation Prediction}
  \label{fig:learning_curve_rotation}
\end{figure}

The images fed to the model are scaled to a uniform size of $50\cdot 50$. As model architecture, a $4$-layer CNN (channels: $64$, $128$, $256$, $512$; kernel sizes: $5$, $2$, $2$, $3$; max-pooling: $1$, $2$, $2$, $1$) followed by an $3$-layer MLP (cell counts: $4096$, $2048$, $1024$). During training, an SGD with a learning rate of $0.02$, a weight decay of $10^{-5}$ and a momentum of $0.95$ was used. The initial learning rate was decreased after every epoch by a fixed gamma of $0.99$. MSE was used as loss metric. For measuring accuracy, every prediction which deviates from the ground-truth angle not more than $5$ degrees was considered correct. In order to verify how the model utilizes the template input, the experiment was conducted twice with identical parameters (once with the image snippets as input only, once with additional symbol templates as input). The template-assisted model outperformed the model that only image snippets as input(see fig.~\ref{fig:learning_curve_rotation}). More precisely, the template-assisted model achieved $53.37\%$ accuracy, while the snippet-only model achieved $46.42\%$ accuracy.

\subsection{Text Recognition}
\begin{figure}[H]
  \centering
  \includegraphics[width=0.6\textwidth]{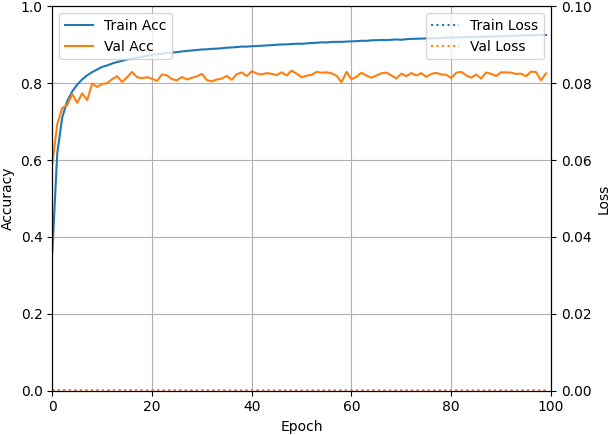}
  \caption{Learning Curve for Text Recognition}
  \label{fig:learning_curve_text_lstm}
\end{figure}

The input images for text recognition are scaled to a height of $50$ while the width is zero-padded to $350$. For this experiments, texts up to a maximum character length of $6$ are considered only. The model consists of a $3$-layer CNN (channels: $64$, $128$, $256$; kernel sizes for all: $5$; max-pooling for all: $2$) followed by a bidirectional LSTM with $2048$ cells and a two-layer CNN of kernel size $1$ and $512$ intermediate channels (there are $97$ character classes). Using an SGD optimizer with a earning rate of $0.8$, a weight decay of $10^{-5}$ and a momentum of $0.9$ as well as statically descreasing learning rate after each epoch with a gamma of $0.99$, the model achieved a character error rate on the validation set of $83.27\%$ (see fig.~\ref{fig:learning_curve_text_lstm}).

\subsection{Segmentation}
\begin{figure}[H]
  \centering
  \includegraphics[width=0.6\textwidth]{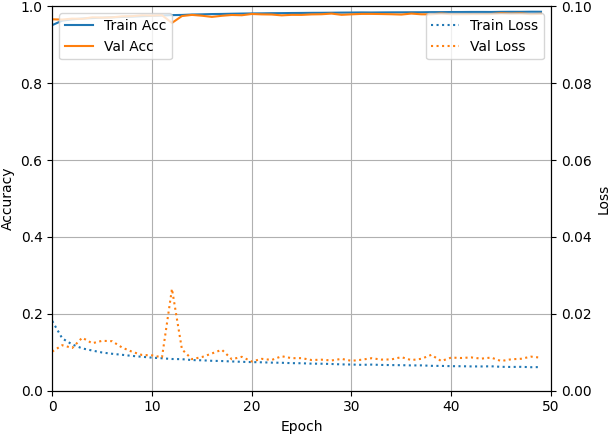}
  \caption{Learning Curve for Binary Segmentation}
  \label{fig:learning_curve_segmentation}
\end{figure}

A U-Net was trained using a patch-size of $256\cdot 256$, and $200$ randomly cropped and randomly rotated patches per training image. A simple SDG optimizer was used with a learning rate of $0.1$, a weight decay of $10^{-5}$ and a batch size of $4$. MSE loss and binary classification accuracy were used both on the pixel level. The model achieved an accuracy of $98.16\%$ on the validation set and simultaneously $98.57\%$ on the training set (see fig.~\ref{fig:learning_curve_segmentation}).

\subsection{Sample Application}
\begin{figure}%[htp]
  \centering
  \subfloat[Raw Image]{%
  \includegraphics[width=0.49\textwidth]{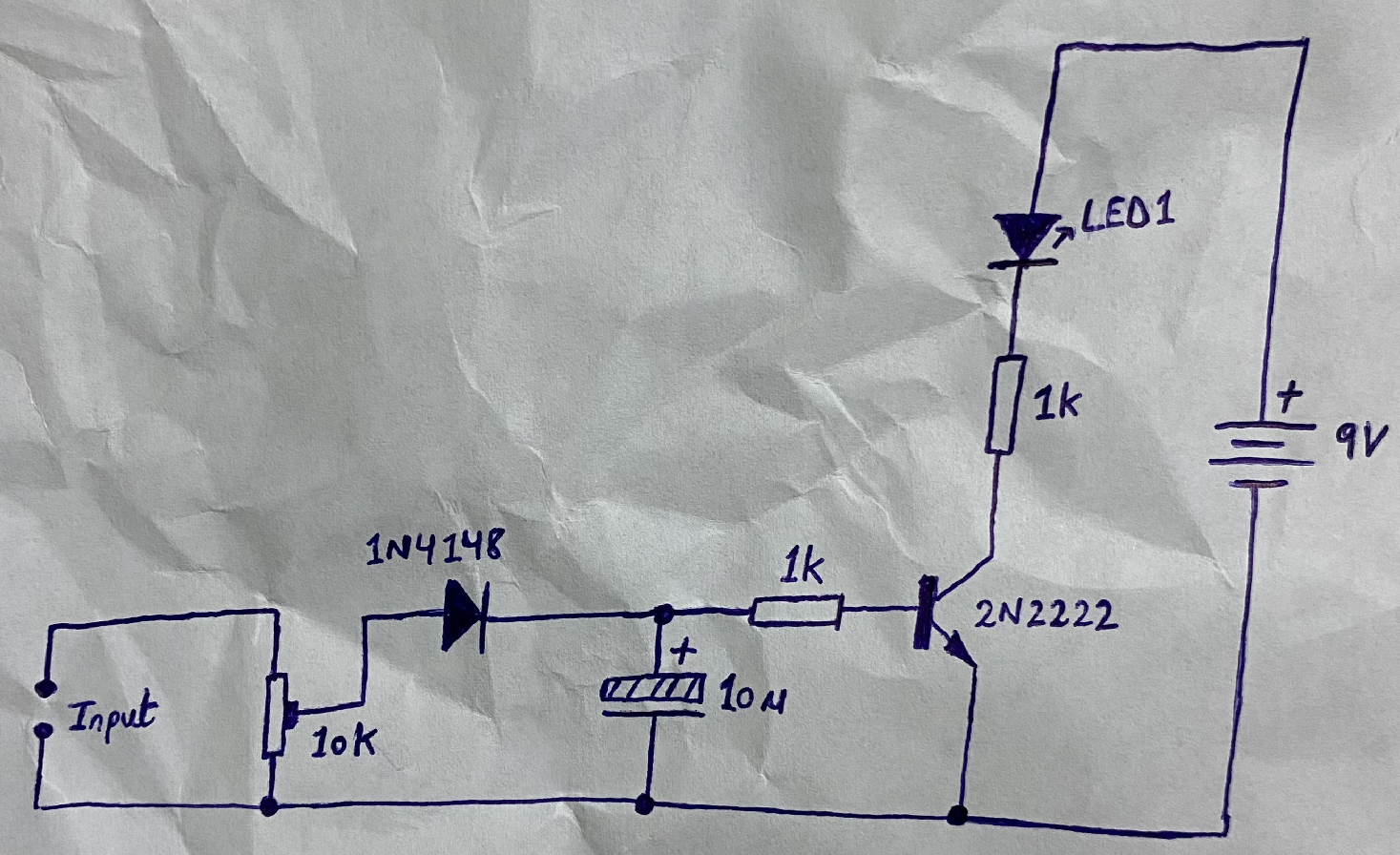}%
  }
  \hfil
  \subfloat[Object Detection]{%
  \includegraphics[width=0.49\textwidth]{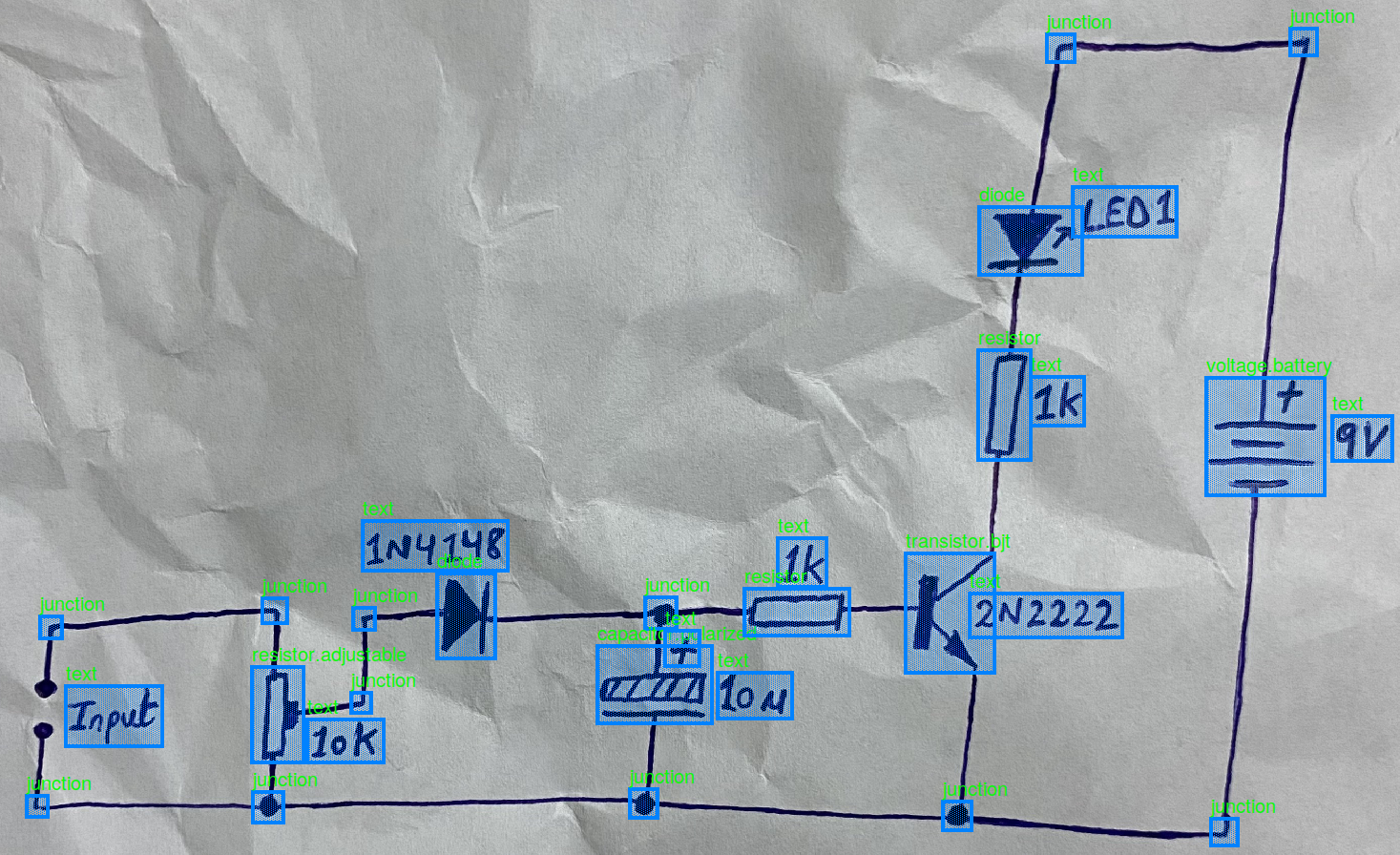}%
  }
  \hfil
  \subfloat[Orientation and Text Recognition]{%
  \includegraphics[width=0.49\textwidth]{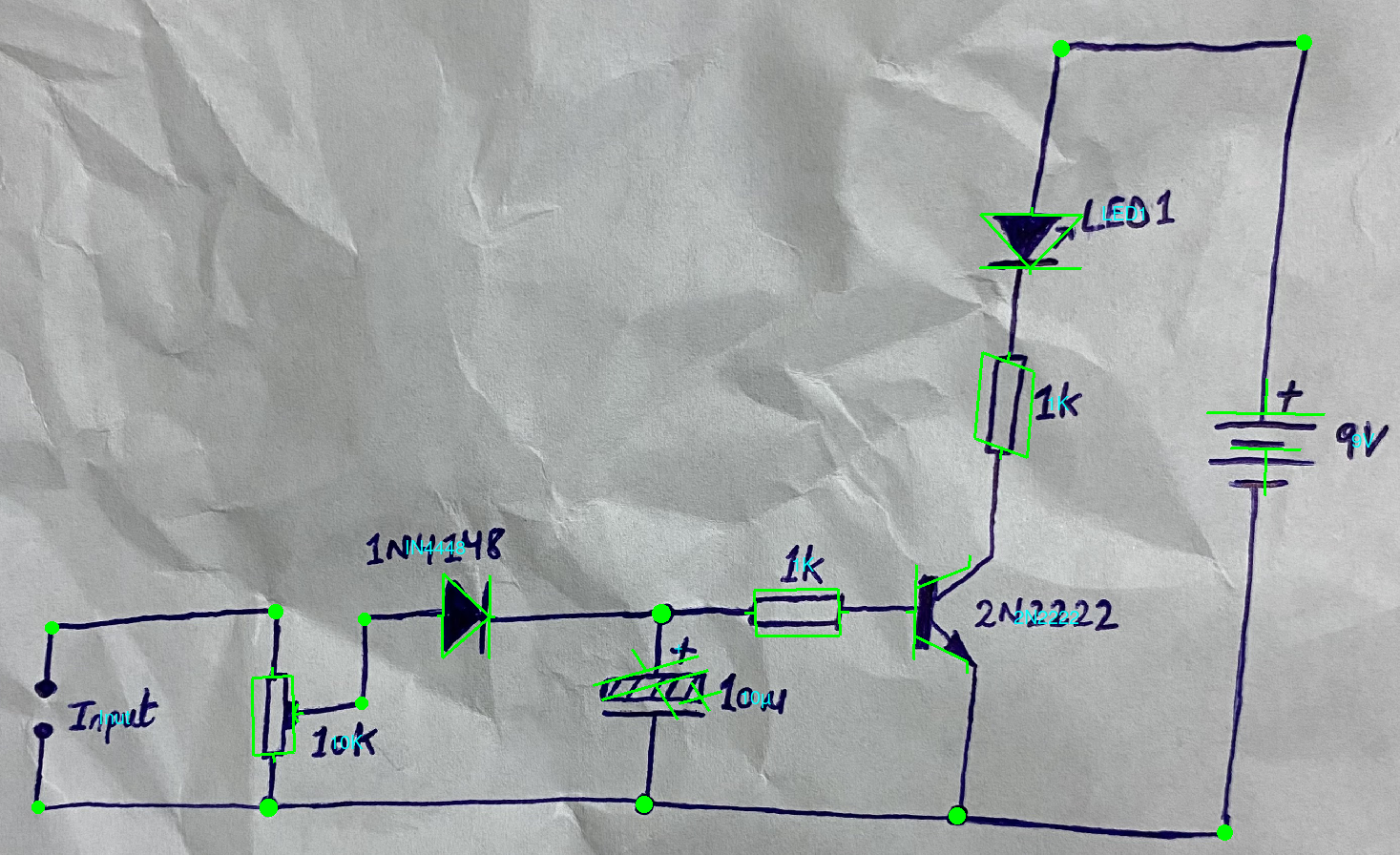}%
  }
  \hfil
  \subfloat[Edge Extraction]{%
  \includegraphics[width=0.49\textwidth]{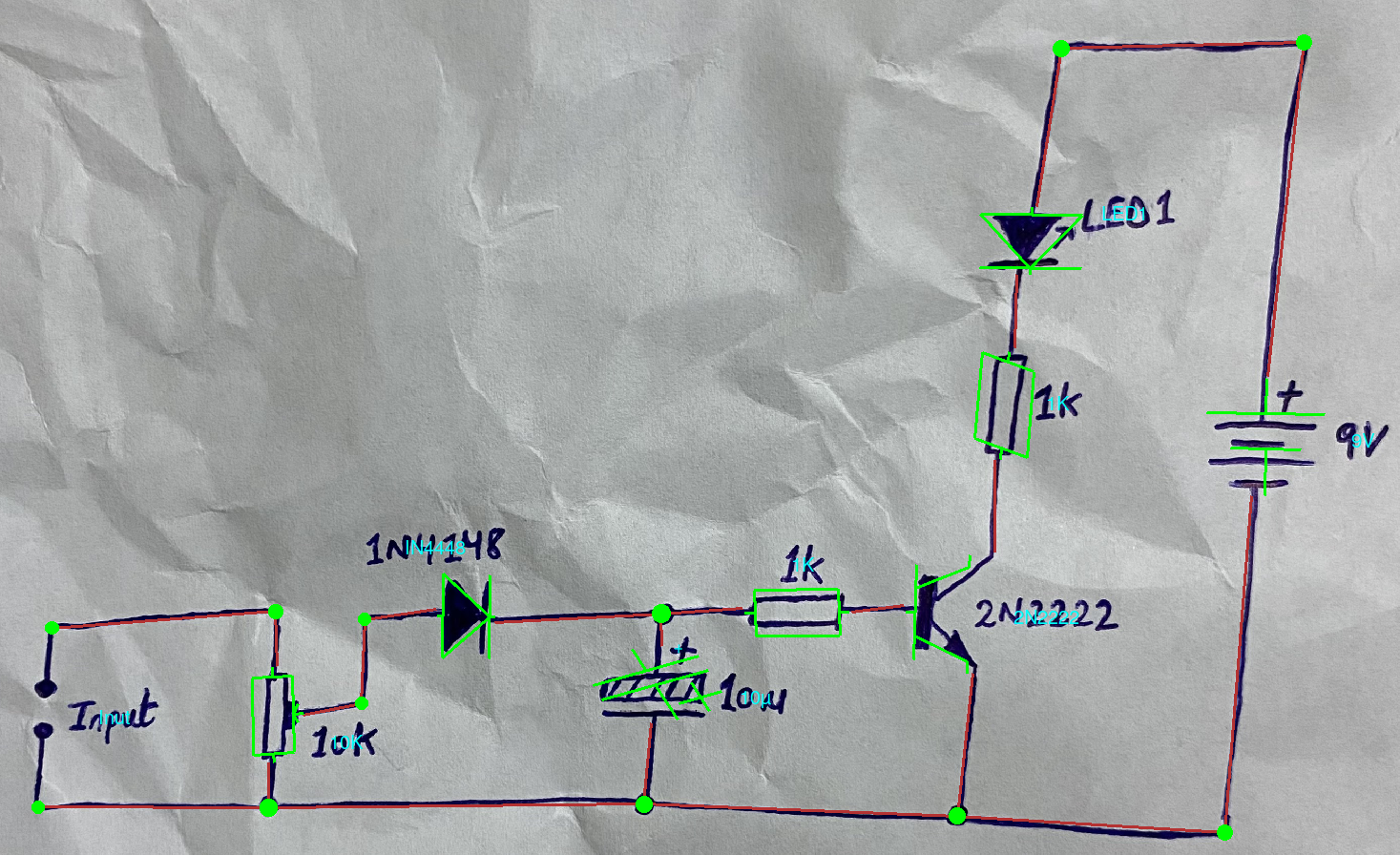}%
  }
  \hfil
  \subfloat[Edge Line Segments]{%
  \includegraphics[width=0.49\textwidth]{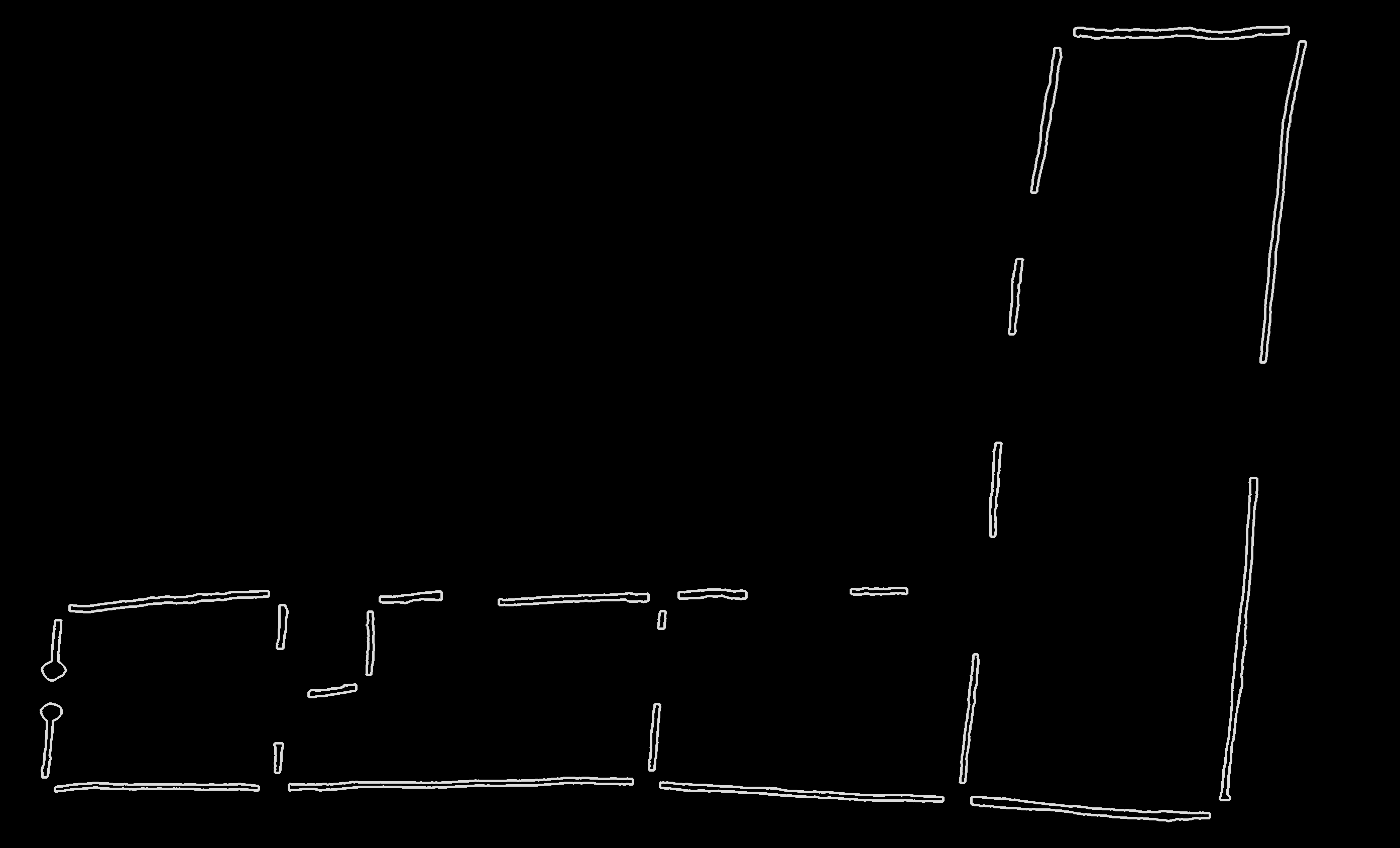}%
  }
  \hfil
  \subfloat[Graph Rectification]{%
  \includegraphics[width=0.49\textwidth]{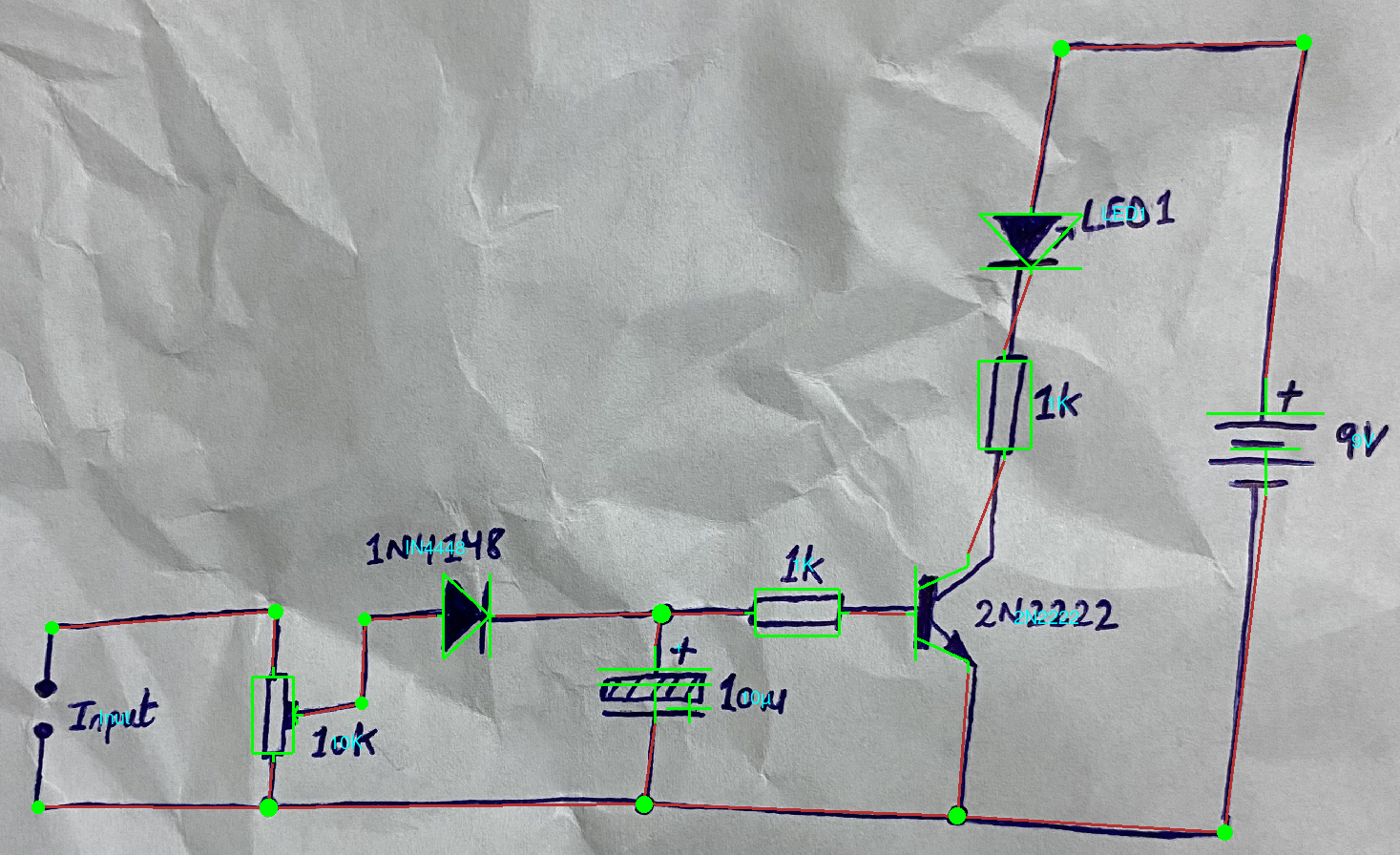}%
  }
  \caption{Application of the Recognition Pipeline to an Sample of the Test Set}
  \label{fig:sample_application}
\end{figure}

\begin{figure}
  \centering
  \includegraphics[width=\textwidth]{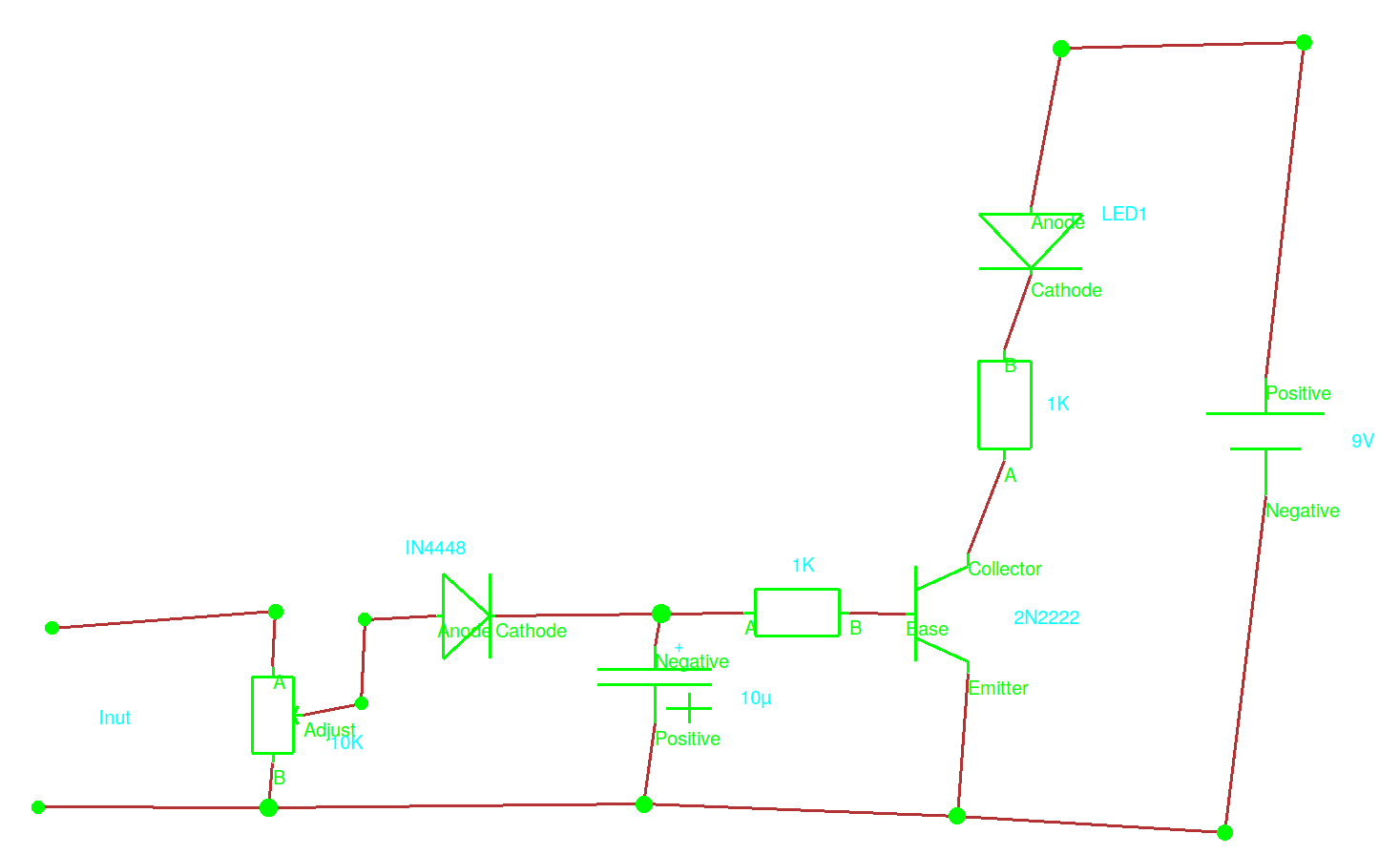}
  \caption{Reconstructed Graph Structure}
  \label{fig:sample_graph}
\end{figure}

In order to validate the entire approach, a sample image from the test set is processed (see fig.~\ref{fig:sample_application}), showing general reconstruction of the graph structure (see fig.~\ref{fig:sample_graph}).

\section{Discussion}

The experiments conducted demonstrate the general viability of the described end-to-end solution. In contrast to most other existing literature, both dataset and code for training and inferencing are publicly available. The modular nature allows to apply the approach both for other symbols, application domains and subsequent processing. Particularly, the orientation prediction which is aided by symbol template inputs and the port assignment aim to be class-agnostic.

\section{Future Work}

Most importantly, the individual tasks like orientation regression and text recognition need to be integrated into a single model with the object detection, thus allowing to utilize the perceptive fields of the intermediate feature maps for better performances. Likewise, a unification of object detector and segmentation models would allow for shorter inference times. Generally, the public dataset can be extended by more drafters and other symbol types.

The obtained graph structures need to investigated for their subsequent usage in the educational domain, i.e. how well they can support students in their learning process or teachers for assessing the student's skills. For that purpose, the graph extraction functionality needs to be integrated into a (e.g. mobile web) application or intelligent tutoring system suitable for students which provides interactive feedback with regard to predefined learning objectives.

\section*{Acknowledgments}
The authors declare that they have no competing interests.

\bibliographystyle{splncs04}
\bibliography{icdar24}

\begin{thebibliography}{10}
\providecommand{\url}[1]{\texttt{#1}}
\providecommand{\urlprefix}{URL }
\providecommand{\doi}[1]{https://doi.org/#1}

\bibitem{icpram23}
Bayer., J., Roy., A., Dengel., A.: Instance segmentation based graph extraction
  for handwritten circuit diagram images. In: Proceedings of the 12th
  International Conference on Pattern Recognition Applications and Methods -
  ICPRAM. pp. 926--931. INSTICC, SciTePress (2023).
  \doi{10.5220/0011752600003411}

\bibitem{bayer2023text}
Bayer, J., Turabi, S.H., Dengel, A.: Text extraction for handwritten circuit
  diagram images. In: International Conference on Document Analysis and
  Recognition. pp. 192--198. Springer (2023)

\bibitem{he2016deep}
He, K., Zhang, X., Ren, S., Sun, J.: Deep residual learning for image
  recognition. In: Proceedings of the IEEE conference on computer vision and
  pattern recognition. pp. 770--778 (2016)

\bibitem{hochreiter1997long}
Hochreiter, S., Schmidhuber, J.: Long short-term memory. Neural computation
  \textbf{9}(8),  1735--1780 (1997)

\bibitem{lakshman2019handwritten}
Lakshman~Naika, R., Dinesh, R., Prabhanjan, S.: Handwritten electric circuit
  diagram recognition: An approach based on finite state machine. Int J Mach
  Learn Comput  \textbf{9},  374--380 (2019)

\bibitem{rabbani2016hand}
Rabbani, M., Khoshkangini, R., Nagendraswamy, H., Conti, M.: Hand drawn optical
  circuit recognition. Procedia Computer Science  \textbf{84},  41--48 (2016)

\bibitem{rachala2022hand}
Rachala, R.R., Panicker, M.R.: Hand-drawn electrical circuit recognition using
  object detection and node recognition. SN Computer Science  \textbf{3}(3),
  ~244 (2022)

\bibitem{ren2015faster}
Ren, S., He, K., Girshick, R., Sun, J.: Faster r-cnn: Towards real-time object
  detection with region proposal networks. Advances in neural information
  processing systems  \textbf{28} (2015)

\bibitem{ronneberger2015u}
Ronneberger, O., Fischer, P., Brox, T.: U-net: Convolutional networks for
  biomedical image segmentation. In: Medical Image Computing and
  Computer-Assisted Intervention--MICCAI 2015: 18th International Conference,
  Munich, Germany, October 5-9, 2015, Proceedings, Part III 18. pp. 234--241.
  Springer (2015)

\bibitem{thoma2021public}
Thoma, F., Bayer, J., Li, Y., Dengel, A.: A public ground-truth dataset for
  handwritten circuit diagram images. In: International Conference on Document
  Analysis and Recognition. pp. 20--27. Springer (2021)

\end{thebibliography}

\end{document}